\documentclass[11pt]{article}
\usepackage{coling2020}
\colingfinalcopy
\usepackage{times}
\usepackage{url}
\usepackage{latexsym}
\usepackage{booktabs}
\usepackage{enumitem}
\usepackage{graphicx}
\usepackage{tabularx}
\usepackage{comment}
\usepackage{todonotes}
\usepackage{booktabs}

\usepackage{pgfplots}
\pgfplotsset{compat=1.14}
\usepackage{pgfplotstable}

\definecolor{aqua}{rgb}{0.0, 1.0, 1.0}

\newlength\graphheight
\newlength\graphwidth
\title{Measuring Uncertainty 
in Translation Quality Evaluation (TQE)}
\author{ Serge Gladkoff$^1$  Irina Sorokina$^{1,2}$     Lifeng Han$^{3*}$  \and Alexandra Alekseeva$^{4}$‡  \\
          $^1$ Logrus Global LLC \\ $^2$ Tver State University \\
$^3$ ADAPT Centre, Dublin City University\\ $^4$ ROKO Labs (‡ deceased) \\
\{serge.gladkoff, irina.sorokina\}@logrusglobal.com \\ \textit{* corresponding}: lifeng.han3@mail.dcu.ie \\
}

\date{}

\begin{document}
\maketitle

\textbf{Abstract:} From both human translators (HT) and machine translation (MT) researchers' point of view, translation quality evaluation (TQE) is an essential task. Translation service providers (TSPs) have to deliver large volumes of translations which meet customer specifications with harsh constraints of required quality level in tight time-frames and costs. MT researchers strive to make their models better, which also requires reliable quality evaluation. While automatic machine translation evaluation (MTE) metrics and quality estimation (QE) tools are widely available and easy to access, existing automated tools are not good enough, and human assessment from professional translators (HAP) are often chosen as the golden standard \cite{han-etal-2021-TQA}.
Human evaluations, however, are often accused of having low reliability and agreement. Is this caused by subjectivity or statistics is at play? How to avoid the entire text to be checked and be more efficient with TQE from cost and efficiency perspectives, and what is the optimal sample size of the translated text, so as to reliably estimate the translation quality of the entire material? This work carries out such a motivated research to correctly estimate the confidence intervals \cite{Brown_etal2001Interval} 
depending on the sample size of translated text, e.g. the amount of words or sentences, that needs to be processed on TQE workflow step for confident and reliable evaluation of overall translation quality.
The methodology we applied for this work is from Bernoulli Statistical Distribution Modelling (BSDM) and Monte Carlo Sampling Analysis (MCSA). %\footnote{*corresponding author: LH}

%to correctly evaluate the overall translation quality.

%\section*{Keywords}

\section{Introduction}
\label{sec:intro}
Machine Translation (MT) is one of the pioneering artificial intelligence (AI) tasks dating back from the 1950s \cite{Weaver1955}. It emphasizes the interaction of Language and Machine, and how machine can learn human languages with cognitive knowledge. Before MT, human translation (HT) of written text and documents has always played an important role in science and literature communication between different language speakers, breaking the language barriers. From both MT and HT perspectives, translation quality evaluation (TQE), sometimes incorrectly referred as translation quality assessment (TQA)\footnote{ISO 9001 distinguishes between ``assessment'' (which relates to the quality management process per se) and evaluation (which is the quality measurement task), so it is incorrect to refer to quality evaluation as ``quality assessment'' \url{https://www.iso.org/iso-9001-quality-management.html}.}, is an important task to reflect how well the source text is translated into the target languages \cite{han-etal-2021-TQA}. 
On the one hand, for low resource language pair scenarios, human translators still play the dominant role in translation production. The translated text and documents can contain unavoidable errors due to personal bias, input efforts, or the training level of the translators. On the other hand, for high resource language pair situation, neural MT (NMT) has achieved remarkable improvement especially on fluency level, compared to conventional rule-based and statistical phrase-based MT models; however, NMT still has ``poisoned cookie'' problem struggling to achieve real human parity, for instance, on adequacy level, meaning preservation, and on idiomatic expression translations \cite{Sag2002MWE,han-etal-2020-multimwe,Google2016MultilingualNMT,han-etal-2020-alphamwe}.
Translation service providers (TSPs) relying on both MT, HT, and human post-editing of MT output (TPE) carry out translation and editing tasks with the high demand and harsh constraints nowadays. Thus, TQE role in this workflow remains to be critical. However, it is tedious, costly, and time-consuming to check through the entire translated text given the huge amount of data TSP and customers process.

One obvious solution, to this point, is to extract a sub-set of the translated text and make a conclusion about the overall translation quality by results of TQE of the sample, which has always been done in real practice. However, one question arises here: \textit{how large the sample size shall be to estimate the overall translation quality of the entire material reliably?} In other words, what is the confidence interval of such evaluation on certain desired confidence level (which is commonly taken as 95\%) with the samples we choose to estimate the overall translation quality?

In this work, we carry out such a motivated experimental investigation on confidence evaluation of translation quality evaluation. To take advantage of statistical modeling techniques, we start with the assumption that error distribution is uniform across the entire material, and minimum unit where the error occurs is one sentence (since errors can be between words, in the form of punctuation and conjugation, etc.). This assumption is the best case scenario which potentially ensures the smallest size of the sample, because if it is not correct we immediately arrive to situation when we need to check more, if not entire text. We start from high quality translation assumption that errors are rare (average error density is as low as 0.07 errors per sentence). Then, on the second stage of experiment, by taking post-editing distance (PED) measurement for evaluation, we prove that such assumption can be applied to the situation where errors are much more frequent, with large number of ``errors'' (edits) per one sentence. We use Bernoulli Statistical Distribution Modeling (BSDM) and Monte Carlo Sampling Analysis (MCSA) to explore the confident level sample size estimation. With this methodology, we expect to reach a conclusion where practical suggestion can be given on confident sample size of translated text/document quality evaluation.

To our knowledge, this is the first study to carry out statistical confidence intervals estimation for MT quality assessment, while in all practical situations, researchers and practitioners very often take it for granted by randomly choosing a sample size from MT outputs to estimate the system quality.

This paper is organized as below: Section \ref{sec:intro} presents the topic of this work, Section \ref{sec:related} covers some related work to our study, Section \ref{sec:method_statistic} introduces the statistical modelling, Section \ref{sec:monte_carlo} presents the Monte Carlo (MC) simulation, Section \ref{sec:monte_carlo_for_ped} carries out confidence estimation for post-editing distance metric using MC simulation, and finally Section  \ref{sec:conclude} concludes the work with discussions and future work.

%\textit{Tips: we can follow the paper structure from ``Machine Translation Quality and Post-Editor Productivity (2016AMTA)'' by first attempt in the research task.}

\section{Related Work}
\label{sec:related}
%\label{sec:method_statistic}
%\label{sec:monte_carlo}
%\label{sec:monte_carlo_for_ped}
%\label{sec:conclude}

Quality estimation and measurement have been explored using various strategies for both human and machine translated text. For instance, human evaluation criteria have included intelligibility, fidelity, 
fluency, adequacy, and comprehension as earlier stage development. These are often carried out based on both source text and candidate translations. Then, editing distance, or post-editing distance are applied to such study by calculating editing steps that are needed to revise the candidate translations towards a correct text sometimes using reference translations. 
Whenever reference translations are available, automatic reference-based evaluation of the candidate translation quality can be measured via many computational metrics including n-gram based simple word matching as well as some linguistic motivated features such as paraphrasing and Part-of-Speech tagging \cite{han-etal-2021-TQA}. 
For automatic post-editing distance based metrics, algorithms are designed to calculate the editing steps of insertion, deletion, substitution and/or word reordering. 
While automatic metrics have been getting very popular due to their low cost and repeatability, and thus easier to perform comparisons to previously published work, the credibility of such automatic evaluations have always been an issue. 
For instance, in contrast to the findings from conversational statistical MT researchers \cite{sanchez-torron-koehn-2016-machine_MT_Post_edit},
very recent research work on Neural MT (MT) by \cite{Zouhar2021NMT_Postediting,google2021human_evaluation_TQA,HanJonesSmeatonBolzoni2021decomposition4mt_MWE} has shown that automatic evaluation metrics such as BLEU do not agree well with professional translators or experts based human evaluations, instead they tend to correlate closely to lower-quality crowd-source based human evaluation.
Another disadvantage for automatic evaluation is that we can not get in-depth view of what kind of errors the candidate translations present in the studied context, except for an overall evaluation score, or segment-level scores, not to mention that most metrics do not even allow for clear interpretation of what does the score exactly mean. Regarding this aspect, professional translators can always do a much better job by giving transparent error analysis and categorization on the candidate translations, such as idiomatic expressions \cite{han-etal-2020-alphamwe,han-etal-2020-multimwe}.

However, another issue arrives at this point, that is how to correctly chose a confident sample from the candidate translation, instead of just take a random sample for granted and try to blindly extrapolate its result to the entire material?
Actually this is not a brand new challenge in natural language processing (NLP) field. 
Having it in mind that randomly chosen samples may contain model bias against a proper evaluation, \cite{prabhu-etal-2019-sampling_classification} proposed an uncertainty sampling approach for \textit{text classification} task, and their statistical models can reduce the bias effectively with smaller size of data in comparison to confessional models;
Similarly, \cite{haertel-etal-2008-assessing} carried out work on how statistical sampling models can help reduce the high cost for Penn Tree-bank annotation while maintaining the higher accuracy;
\cite{nadeem-etal-2020-systematic_sampling} carried out one systematic comparison of several sampling algorithms used for language generation task, including top-\textit{k}, nucleus and tempered sampling, looking into quality-diversity trade-off.  %4team: if it is possible, we can compare our Monte-carlo sampling results to these three or any of them to see which is better in quality-diversity trade-off?
%4team: \url{https://aclanthology.org/search/?q=sampling+}
Sampling method was also applied into confident level evaluation of MT. For instance,  \cite{koehn-2004-statistical-significance}  proposed to use bootstrap re-sampling methods to test the significance level of automatic metric BLEU, but using a fixed number of sentences, i.e. 300. Like many other research work, the chosen number of sentences for evaluation was never explained or justified with any statistical validation. 
In contrast, we carry out statistical sampling modelling to estimate the number of sentences that is confident enough to achieve reliable quality evaluation, which means a better representation and generalization of the overall candidate translations in question, with an confidence-cost tread-off.

%to achieve represenative sample with better generalization...

%I. quality estiomation of MT is important; methods including ... ; how about confidence sample size? 

%II. post-editing has been applied in MT assessment;
%There are researchers who carried out experimental investigation on the correlation between post-editing efforts that are needed to correct the MT output and BLEU score variations

%PED used for SMT \cite{sanchez-torron-koehn-2016-machine_MT_Post_edit} and NMT quality measurement <Neural Machine Translation Quality and Post-Editing Performance> 2016amta 2021emnlp. \cite{laubli-etal-2019-post_edit_NMT,Zouhar2021NMT_Postediting}

%III. sampling methods have been used in NLP;

%\cite{Zouhar2021NMT_Postediting} discussed the post-editing relation to NMT quality, and BLEU apparently not an choice.

\begin{comment}
...
MT significance testing
transaltionese
quality estimation
post-editing distance
\end{comment}

\section{Statistical Modelling}
\label{sec:method_statistic}
%\label{sec:monte_carlo}
%\label{sec:monte_carlo_for_ped}
%\label{sec:conclude}

\subsection{Study Setup}
\label{subsec:study_setup_for_sta_model}
To carry out a statistical modeling of our research questions under study, i.e. \textit{how to confidently choose a sample size of translation outputs to estimate the overall translation quality, from either HT or MT}, we setup the following initial assumptions: 
%We have the following assumptions for our methods to be carried out:
\begin{itemize}
\item Translation errors belong to several independent categories
\item Errors of one category are independent from each other.
\item Errors of one category occur in text randomly.
\item The smallest unit of text where language error occurs is a sentence (in other words, error can be between words, but cannot be between sentences).
\item The resulting error distribution is a superposition of distribution of errors of different categories.
\end{itemize}

Then, if we further project these assumptions into simpler mathematical notes as bellow, it meets definition of Bernoulli trial \cite{Brown_etal2001Interval,Agresti_Coull1998_approximate}:
\begin{itemize}
\item Errors of certain type (category) either present in a sentence, or not.
\item Errors are independent from each other.
\item The probability of errors is the same.
\end{itemize}

Generally speaking, we cannot always assume that the distribution of errors in the text is uniform, but for the purpose of this exercise it is a reasonable assumption because it is the best possible scenario. 
The reason is that if errors themselves are not normally distributed across the text, then errors are not distributed uniformly across the material, and the sample size should be much larger to capture clusters of errors, or even 100\% review of the material may be required to fund such clusters. Therefore, the assumption that errors are uniformly distributed is the best scenario, which allows for the smallest sample size possible.

While this assumption may be questioned, in our experimental evaluations, it proves that the starting value of random seed does not affect the overall model behavior and its solutions. For high quality human translation, these assumption can be actually very reasonable. Regarding MT outputs and less than premium quality translation scenarios, where many errors per one sentence is not something unusual, we will deal with such more error-prone materials separately in later sections of this paper (Section \ref{sec:monte_carlo}). %, because simplified model does not apply to them.

\subsection{Bernoulli Distribution}
\label{subsec:bernoulli}
In Bernoulli statistical distribution, when the sample size $n$ is significantly smaller than the overall population $N$, the standard derivation of sample measurement falls into the following formula: 

 \[ \sigma = \sqrt{\frac{p\cdot(1-p)}{n}} \]

\noindent where \textit{p} is the probability estimation of an event under study. The confidence interval $CI$, using the Wald interval \cite{Newcombe-book2012confidence}), will be:

\[ CI = p \pm \Delta \]

\noindent where $\Delta$ is the product of standard deviation and factor 1.96 (when confidence level 95\% is chosen) \cite{Agresti_Coull1998_approximate}.

\[ \Delta = 1.96\cdot \sigma =1.96 \cdot \sqrt{\frac{p\cdot(1-p)}{n}} \]

%The confidence interval (Wald interval \cite{Newcombe-book2012confidence,Agresti_Coull1998_approximate}) will be:

% \[ CI = p \pm \Delta \]

When the sample size $n$ is comparable to the population size $N$, e.g. in a smaller translation evaluation project for our study, the standard deviation is calculated as bellow and the $\Delta$ value updates correspondingly:

\[ \sigma = \sqrt{\frac{p\cdot(1-p)}{n}\cdot\frac{N-n}{N-1}} \]\\

%We carried out a case study when $p$ and $\Delta$ values are chosen as 0.07 and 0.02 (confidence interval 0.07 ± 0.02, i.e. from 0.05 to 0.09), the estimated sample size $n$ will result in 625 sentences to be checked. 

Let's come back to our study, with this even distribution assumption of each sentence regarding translation errors, having error probability $p$ with value 1 and no error probability  $1 - p$ with value 0,  each sentence represents a random variable in the modelling. 
%Come back to our study, regarding the  population of sentences, every sentence represent a random variable which takes the value 1 (has error) with probability $p$ and value 0 (no error) with probability $1 - p$.
This forms a Binomial distribution \textit{B(n, p)} where \textit{n} is the number of sentences.

%Surly, there can be several errors of one type in a sentence; looking for them is making a fixed number \textit{n} of statistically independent Bernoulli trials, each with probability of \textit{p}, which comprises a Binomial distribution.
%We therefore adopt a model where errors of one type have Binomial distribution \textit{B(n, p)}.

\subsection{Case Studies}
\label{subsec_case_study_of_Statistical_Modelling}
We present case studies using both high quality translation text and low quality one. 
We first carried out a case study using high quality translation. Statistics from language service providers \footnote{for instance, \url{https://logrusglobal.com/}} shows that the average length of English sentence is 17 words; there are about 250 words on standard page; we therefore can assume that there are 15 English sentences on a standard page.
Let's assume that there is very high quality translation document, where there are no more than one error per page (one error per 15 sentences); then error density $p = 0.07$.  And, if we set $\Delta$ value as 0.02. Thus the confidence interval falls into 0.07 ± 0.02, i.e. from 0.05 to 0.09, which is already wide interval. If we use a confidence level 95\%,
 we have the following recommended number of sentences to check as derived from the formula mentioned earlier:

%with error density 0.07 ($p$ value) and $\Delta$ value chosen as 0.02. Thus the confidence interval falls into 0.07 ± 0.02, i.e. from 0.05 to 0.09, which is already wide interval. Then the corresponding confident sample size  $n$ is inferred to be 625 sentences.

%In this case we can derive the required volume of text to check from the formula above, and it will look like this:\\

\[ n = {\frac{1.96^2 \cdot p \cdot (1-p)}{\Delta^2}} \]

\[ n = {\frac{1.96^2\cdot0.07\cdot(1-0.07)}{0.02^2}} = 625 \]

\begin{figure}%[H]
\begin{center}
\includegraphics[width=0.9\textwidth]{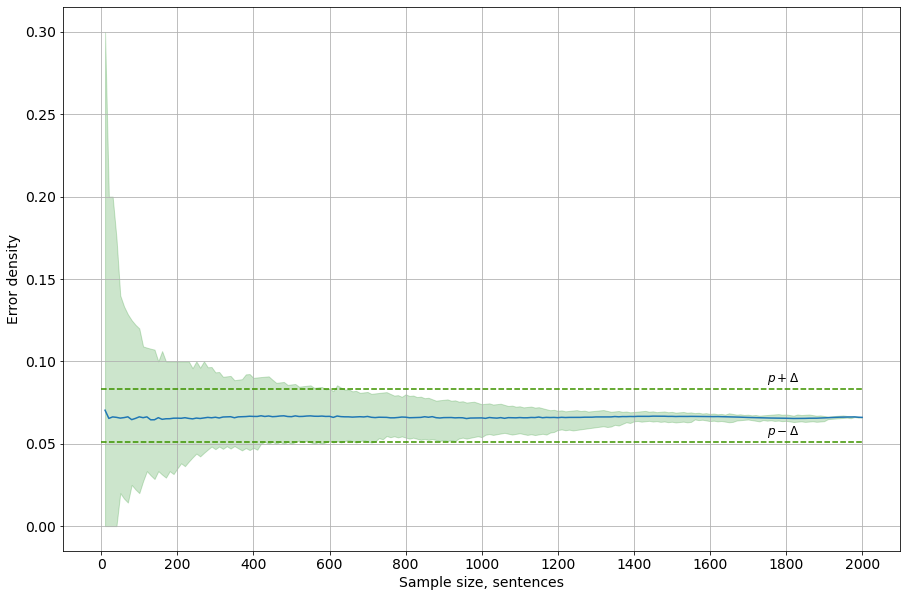}
\end{center}
\caption{A 95\% confidence level credible interval for sample size from 100 to 2000 sentences, for high quality translation with error density 0.07. $\Delta$ is shown for sample size 625 sentences (42 pages).}
\label{fig:high_MT_625_sent}
\end{figure}

%Let's get back to our example where average error density $p$ is 0.07. We probably want to measure with error not more than $\Delta$ = 0.02, then our measurement will be within confidence interval $0.07 \pm 0.02$, i.e. from 0.05 to 0.09 (already wide interval). 
%If we want to achieve certain error tolerance with confidence 95\%, we need to know what is the error density in the document; let assume that this is very high quality document, where there are no more than one error per page (one error per 15 sentences); then $p = 0.07$ and we have the following recommended number of sentences to check:

\noindent which requires us to check 625 sentences, or 42 pages, or 10,000 words (all these numbers are rounded) as demonstrated in Figure \ref{fig:high_MT_625_sent}.
If we want to achieve measurement accuracy with confidence interval $0.07 \pm 0.01$ then we will have to check 2500 sentences, or 166 pages, or 42,000 words.
If we check 10 pages the confidence interval will be 0.04, and the measured result will be $0.07 \pm 0.04$ (from 0.03 to 0.11), and if we check only 1 page: $0.07 \pm 0.12$ (from 0 to 0.19).
If we try to check only one sentence even of high quality translation, we really can't say anything about the quality because the error probability of such measurement will be $0.07 \pm 0.5$ (from 0 to 0.57) which is too high.

\renewcommand{\figurename}{Fig.}

\begin{figure}%[H]
\begin{center}
\includegraphics[width=0.9\textwidth]{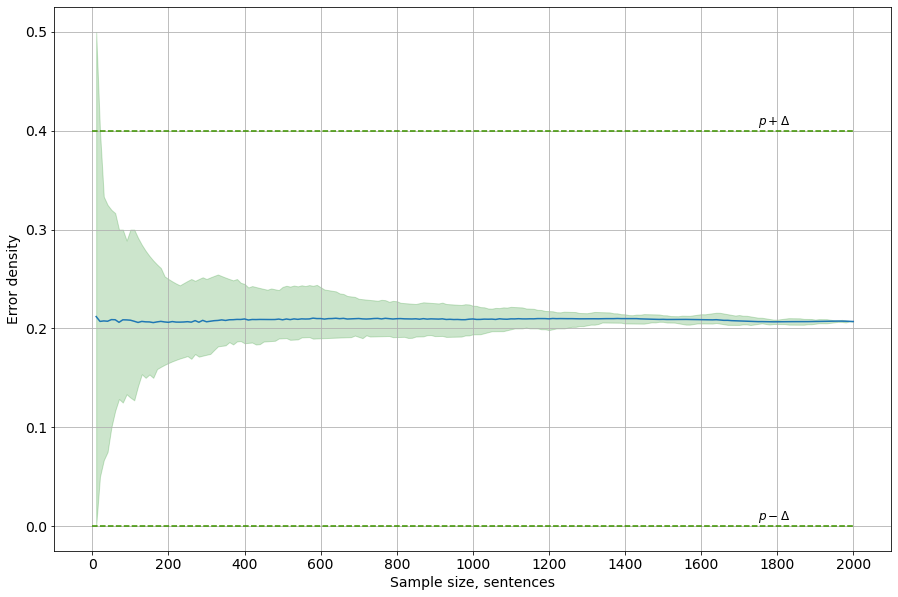}
\end{center}
\caption{A 95\% confidence level credible interval for sample size from 100 to 2000 sentences, for low quality (machine) translation with error density 0.2. $\Delta$ is shown for sample size 15 sentences (1 page).}
\label{fig:low_MT_15_sent}
\end{figure}

Then let's have a look into examples with lower quality translation text, for instance, an error density of 0.2, i.e., one error for every 5 sentences. 
The confidence interval of checking only one page will be $\pm 0.2$: the measurement confidence will be $0.2 \pm 0.2$ and the actual value can be anywhere from 0 to 0.4 as shown in Figure \ref{fig:low_MT_15_sent}. As we can see, the one page measurement in this situation is not reliable at all.

\renewcommand{\figurename}{Fig.}

%\label{fig:high_MT_625_sent}
%\label{fig:low_MT_15_sent}

As we mentioned in Section \ref{subsec:bernoulli}, the formulas above only apply for the cases when the sample size is much smaller than the overall population size.  If sample size $n$ is comparable to the population size $N$, the standard deviation of such measurement in the case of high quality translation evaluation will be differently calculated as:

\[ \Delta = 1.96\cdot\sqrt{\frac{p\cdot(1-p)}{n}\cdot\frac{N-n}{N-1}} \]

Practicality of this can be illustrated with the following example: suppose that we have a translation quality evaluation (TQE) scorecard measuring the quality of 1000 words job by doing TQE on 400 words sample checked by reviewer, who found 2 minor errors. Then measured error density is 2 errors per 23.5 sentences, or 0.085 errors per sentence.
%Practicality of this can be illustrated with the following example: suppose that we have a translation quality evaluation (TQE) scorecard measuring the quality of 1000 words sample, and reviewer checked 400 words and found 2 minor errors. Then measured error density is 2 errors per 23.5 sentences, or 0.085 errors per sentence.

The 95\% confidence interval for such measurement will be $\pm\Delta$, where:

\[ \Delta = 1.96\cdot\sqrt{\frac{0.085\cdot(1-0.085)}{23.5}\cdot\frac{58.82-23.5}{58.82-1}} = 0.088 \]

Therefore, with such a scorecard measurement we can only say that the measured error density is 0.085 $\pm$ 0.088, i.e. with 95\% confidence the actual error density is in the range [0.034 - 0.173]. 
Of course, we can check all 1000 words of this job, and then we will be 100\% certain about the number of errors in this job (assuming that the reviewer is reliable, but that’s another story), but then the measurement confidence interval will relate to the generalized conclusion about the work of this supplier in general, on population size much larger than 1000 words.
%If we check all 1000 words of this job, we will be 100\% certain about the number of errors in this job, but then the measurement confidence interval will relate to the generalized conclusion about work of this supplier.  % and the first formula would apply.\\

%For lower quality texts where error density is 0.2 (one error for every 5 sentences) confidence interval of checking only one page will be $\pm 0.2$: the measurement confidence will be $0.2 \pm 0.2$ (actual value can be anywhere from 0 to 0.4). As you can see, the one page measurement like that is not reliable at all.

%Average length of English sentence is 17 words; there are about 250 words on standard page; we therefore can assume that there are 15 English sentences on a standard page; therefore, if error density is 0.07 and required $\Delta$ = 0.02 with confidence level 95\% the numbers above require us to check 625 sentences, or 42 pages, or 10,000 words (all these numbers are rounded).\\

%For lower quality texts where error density is 0.2 (one error for every 5 sentences) confidence interval of checking only one page will be $\pm 0.2$: the measurement confidence will be $0.2 \pm 0.2$ (actual value can be anywhere from 0 to 0.4). As you can see, the one page measurement like that is not reliable at all.

%If we try to check only one sentence even of high quality translation, we also really can't say anything about the quality because the error of measurement will be $0.07 \pm 0.5$ (from 0 to 0.57).

In summary, we have the following findings from Bernoulli Distribution modeling for the case studies:

\begin{enumerate}

\item If the sample size is much smaller than the entire material, the correct sample size for the measurement does NOT depend on the size of the material, but does depend on the error density in the material and sample size. This case applies to continuous delivery cases, and spot checks on them, or very large projects.

\item If the sample size is comparable with the size of the entire material, the correct sample size for the measurement also depends on the size of the material as well as on the error density in the material and sample size. This case applies to small jobs. 

\item The more precisely we want to measure, the more we need to check.
\end{enumerate}

However,  in reality, the formulas for Bernoulli distribution will not always apply successfully because that the error distribution in the text is sometimes not uniform. 
In order to see what is happening with reliability of quality evaluation in more challenging scenarios, in the next section, we conduct numerical Monte-Carlo sampling experiment to examine how our theory comes to practice and then introduce real life complications.

\section{Monte Carlo Simulation (MSC)}
\label{sec:monte_carlo}
%\label{sec:monte_carlo_for_ped}
%\label{sec:conclude}

Monte Carlo Simulation (MCS) was first developed, in the last century when computer machine was still newborn, for a gambling situation when various outcomes can't be easily predicted or determined due to many random variables. MCS assigns some variables from a task that have certain uncertainty with random values, then runs the model to predict a result, which process is subsequently repeated with a great number of times. Finally, estimation will be achieved by averaging the large number of results.
These results can also be used to assess statistical properties of error distribution in the samples. The repeated process is also called sampling in many applications, such as in our translation quality evaluation task under study \cite{Goodfellow-et-al-2016} (\textit{Chap. 17}).

To conduct a numerical experiment we need to take certain parameters of the translation error distribution in a sample of given size.

It is intuitively clear that, the number of errors in one sample will be different each time when the sample is generated simply because errors are distributed randomly throughout the text. This includes the situations of 100\% correct judgments of errors and no false positives.

It is less obvious that the parameters of error distribution  in the sample are different from parameters of error distribution in the entire material ( \textit{aka} ``population''). 
A degenerated case of minimal sample (e.g. one sentence) helps to understand this: if error density in population is 5\% and we only analyze one sentence, with 95\% probability there will be no errors in that sentence, however it is 100\% sure that there are errors in entire material.

The estimation of rare event probabilities are best analyzed with MCS method, sometimes it is probably the only way to handle such phenomena. 
As we mentioned, MCS relies on repeated random sampling to obtain numerical results for random variables where models are not available analytically. 
In our study, it is the case of translation error in the text, with many types of it which interact in a complex way, and the distribution of these errors is not uniform due to the text having a structure of unequal content. The possible reasons for this include that different people have worked on the entire text, and a plethora of other sophisticated reasons.
%), possibility of different people working on the entire text and a plethora of other sophisticated reasons.\\

Our initial experiment will not be dealing with the complexity of many types of errors but examines the simplified model described in earlier section (Section \ref{sec:method_statistic}).
Correspondingly, our numerical MCS experiment to assess parameters of this distribution is described as  follows:

\begin{enumerate}
\item We take a sample size N = 2000 for repeated process.% random samples of given size from entire material.

\item We generate the random distribution of errors in the entire ``population'' of all the sentences of the material.

\item The number of errors found in these samples represent error distribution of a total number of errors in a sample.

\item We use the large number of sampled data to estimate the entire collection of materials, the error distribution, mean and confidence intervals of such distribution.
%When we form 2000 samples and analyze them we will be able to find out what is the error distribution in a sample of given size, and determine its standard error, mean and confidence interval of that distribution, and therefore will know how reliable is one measurement, because one measurement is only one data point and we have studied the entire collection of such measurements.

\item We take the same error density assumption of 0.07 as in the earlier section (Section \ref{subsec_case_study_of_Statistical_Modelling}).

\end{enumerate}

After, we will be able to assign different values of 
%Later we will be able to change values of 
these parameters and see how this affects the corresponding results.

%As we generated the ``population'' (the data set representing the text with 2000 ``sentences'' and randomly distributed errors in them) we need to generate large number of samples. The number of samples should be large enough to reliably apply statistics to rare random variable. Our chosen number of samples in this experiment is 2000.\\

Now that we have 2000 samples of text to be evaluated, we can look into the parameters of error distribution in such samples, based on a large data-set. 
We can calculate the mean of a number of errors in a sample, by creating a list of 2000 elements with one value - the number of errors found in a sample set.

\begin{figure}%[H]
\begin{center}
\includegraphics[width=0.66\textwidth]{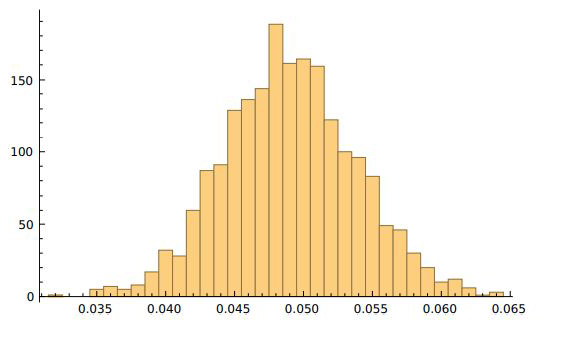}
\end{center}
\caption{Distribution histogram of error number in samples, with sample size of 1000 sentences.}
\label{f_lenratio_trx}
\end{figure}

When this is done we have a 2000 elements array of error numbers in 2000 samples, and we can plot the histogram, as shown on Fig. \ref{f_lenratio_trx}. 
This histogram is a discreet representation of probability density function of distribution of error numbers across the samples. 
It is not a surprise that the distribution is a Bell curve; regardless of whether errors in the entire population are uniformly distributed or not, the number of errors in the sample is normally distributed, provided the number of samples is sufficiently large, according to the Central Limit Theorem.

\renewcommand{\figurename}{Fig.}

\begin{figure}%[H]
\begin{center}
\includegraphics[width=0.66\textwidth]{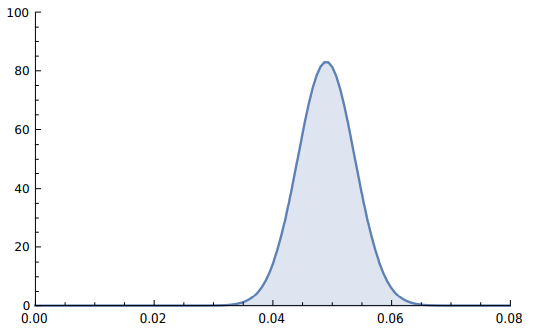}
\end{center}
\caption{Reconstructed probability density function of distribution of error number in samples of 1000 sentences.}
\label{f_lenratio_trx_recon}
\end{figure}

The parameters of this distribution can be reconstructed with \textit{norm.fit} function of \textbf{scipy.stats} \footnote{\url{https://docs.scipy.org/doc/scipy/reference/stats.html}} programming package, as in Figure \ref{f_lenratio_trx_recon}.

\begin{figure}%[H]
\begin{center}
\includegraphics[width=0.5\textwidth]{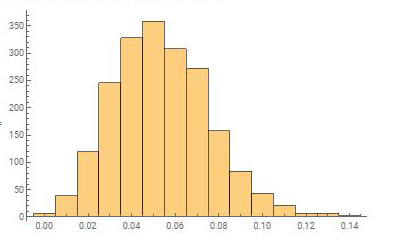}
\end{center}
\caption{Distribution histogram of error number in samples, with sample size of 100 sentences.}
\label{f_lenratio_trx_100}
\end{figure}

Now, if we take a much smaller (10 times smaller) sample size of 100 sentences we will get a much wider bell, as in Figure \ref{f_lenratio_trx_100}.

\begin{figure}%[H]
\begin{center}
\includegraphics[width=0.66\textwidth]{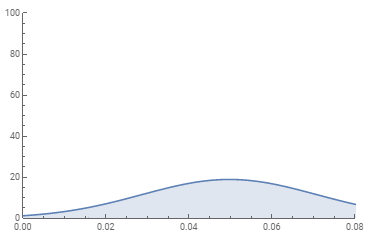}
\end{center}
\caption{Reconstructed probability density function of distribution of error number in samples of 100 sentences.}
\label{f_lenratio_trx_100_recon}
\end{figure}

\noindent which naturally corresponds to wider probability distribution, in Figure \ref{f_lenratio_trx_100_recon}.

If we run a computational experiment to plot 95\% confidence level credible interval for sample sizes from 10 to 2000 sentences, such an experiment will produce results very close to the formulas in previous section (Section \ref{subsec_case_study_of_Statistical_Modelling}), as shown on Figure \ref{fig:high_MT_625_sent}. %. 5:\\
%\label{fig:high_MT_625_sent}
%\label{fig:low_MT_15_sent}

In the next section, we will demonstrate how MCS can be applied to the confidence estimation on language quality metrics (LQMs), such as post-editing distance.

\section{MCS for Post-editing Distance (PED) Confidence Estimation}
\label{sec:monte_carlo_for_ped}
%\label{sec:conclude}

As we mentioned in Section \ref{sec:intro}, in this section, we investigate how MCS can be applied to simulate translation quality evaluation and confidence estimation task. 

Due to the lack of reliable and precise automatic language quality evaluation (LQE) metrics, currently Post-Editing Distance (PED) score remains one of the popular measurement. The PED score is often tracked on segment level, in comparison to document or system level, to examine how good the MT output is in comparison to human edited final translation.

We further conduct analysis of confidence interval (CI, or $\Delta$) for average PED score depending on the sample size.

The absolute PED score is the minimal number of deletion and insertion operation/editing steps form initial candidate translation to post-edited translation text, divided by the length of source sentence. Because the number editing steps can be larger than the number of words in the source sentence, the absolute PED score can be greater than 1.

\begin{figure}%[H]
\begin{center}
\includegraphics[width=0.8\textwidth]{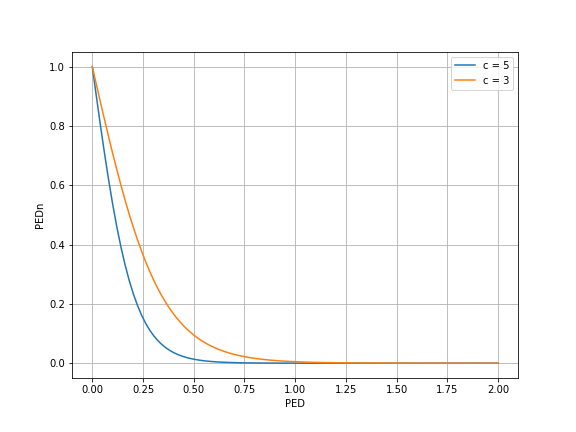}
\end{center}
\caption{Relationship between plain PED and normalized PEDn.}
\label{fig:ped_to_pedn}
\end{figure}

However, since we design to compare the PED with vector similarity, the absolute PED score needs to be normalized to a [0, 1] range. 
%Since we are going somewhere down the road to compare PED to vector similarity, we need to normalize absolute PED to [0, 1] range.\\
We propose a normalization function of PED (represented as PEDn) as bellow: % that the best way to normalize absolute PED is the following function:

\[ PEDn = 1 - tanh (c \cdot PED) \] 

\noindent where \textit{c} is a parameter defining the value of PED which brings the value of normalized PED to [0, 1], as shown in Figure \ref{fig:ped_to_pedn}.

\renewcommand{\figurename}{Fig.}

%Normalized PEDn is designed to be comparable with vector similarity value which should be in the range of [0, 1] where 1 means sentences are identical.\\

Using the MCS methods we introduced in the last section (Section \ref{sec:monte_carlo}), we conduct numerical experiment to estimate confidence interval (Delta) for average normalized measurement PEDn as function of sample size (number of sentences).

This allows us to understand the number of sentences to calculate average PEDn from samples with confidence interval $\pm \Delta$ value.

\begin{figure}%[H]
\begin{center}
\includegraphics[width=0.8\textwidth]{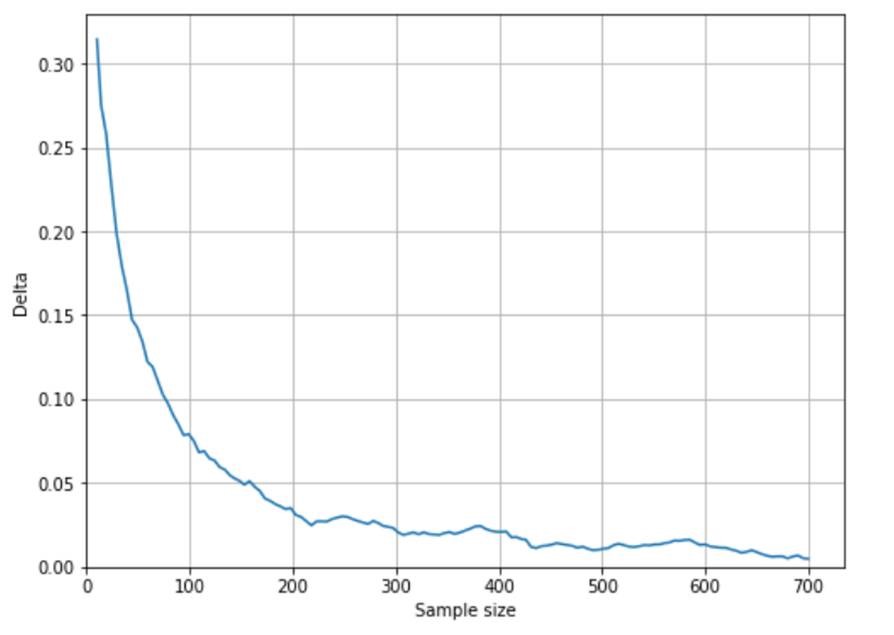}
\end{center}
\caption{Delta as function of sample size.}
\label{fig:delta_as_func_of_sample_size}
\end{figure}

\begin{figure}%[H]
\begin{center}
\includegraphics[width=0.8\textwidth]{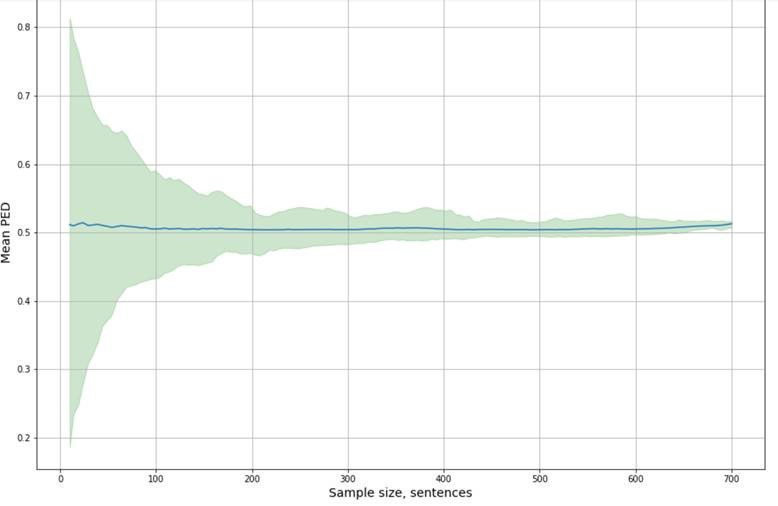}
\end{center}
\caption{A 95\% confidence level credible interval for mean PEDn.}
\label{fig:95_CI_for_mean_pedn}
\end{figure}

First we simulate Delta (half of confidence interval) as function of the sample size, as in Figure \ref{fig:delta_as_func_of_sample_size}. Then the confidence interval is shown in Figure \ref{fig:95_CI_for_mean_pedn} which graphic indicates that confidence interval starts going out of control when the sample size is less than 200 sentences (approximately 3500 words).

\renewcommand{\figurename}{Fig.}

\renewcommand{\figurename}{Fig.}

This means that reliable conclusions can not be reached about the quality of MT proposals based on post-editing sample sizes of 2000 words or less. 
A sample size above 4000 words is recommended to get reasonably confidence interval in a better scenario, for purely statistical reasons, set aside subjective difference of opinions of different editors, content irregularities, and other possible factors.

%...

\section{Discussions and Conclusions}
\label{sec:conclude}

In this article, we proposed the research question on confident sample size of translation documents to estimate the overall material quality, which is a crucial question for both academic research and industrial applications, such as for clients and language service providers. 

We started the experimental investigation of translation quality evaluation (TQE) by assuming that the errors in translated text are evenly distributed, with errors being rare (7 errors per 100 sentences, on average). This assumption is placed as a random seed of our statistical Bernoulli modelling which does not affect the overall model behavior.
%We started the experimental investigation of translation quality evaluation (TQE) by assuming that the errors in translated text are evenly distributed, and no more than one error per translated sentence. This assumption is placed as a random seed of our statistical Bernoulli modelling which does not affect the overall model learning. %. , and the simulation 
%While this assumption may be questioned, in our experimental evaluations, it proves that the starting value of random seed does not affect the overall model learning and the solutions to be reached.  
%High quality translation with error density 0.07 corresponds to a confident sample size 625 sentences. 
To simulate the practical situation where the errors can come from different translators and different types, and span into a different weight across the translated text and documents, we applied the Monte Carlo Simulation analysis, using a sample size of 2000 sentences and 95\% confidence level.
We also applied MCS into confidence estimation of post-editing distance measurement which is currently widely adopted evaluation metric for translation assessment, and gained very valuable findings from empirical investigations 
regarding practical situations when translation quality evaluation (TQE) is deployed.

Furthermore, we suggest that, ideally, a reliability level of analytic sample quality measurement can be added to every analytic TQE scorecard in the form of confidence interval at certain confidence level as one important indicator of the level of certainty of measurement results. In the future work, we plan to compare different sampling methods, as well as apply the confidence estimation model into broader TQE metrics.

\bibliographystyle{coling}
\bibliography{coling2020}

\end{document}